\documentclass[twoside,11pt]{article}
\usepackage{arxiv} 
\usepackage{hyperref}
\usepackage{url}
\usepackage{enumitem}
\usepackage{graphicx}
\usepackage{float}
\usepackage{wrapfig}
\usepackage{siunitx}

\usepackage{amsmath}
\usepackage[capitalize,noabbrev]{cleveref}

\usepackage{microtype}
\usepackage{graphicx}
\usepackage{dsfont}
\usepackage{algorithmic}
\usepackage{mathrsfs}
\usepackage{mathtools}
\usepackage{custom_commands}
\usepackage{enumitem}
\usepackage[noorphans,vskip=0ex]{quoting}
\usepackage{xcolor}
\usepackage{xspace}
\usepackage{bm}
\usepackage{bbm}
\usepackage{etoolbox}
\usepackage{url}
\usepackage{lipsum}
\usepackage{hhline}
\usepackage{float}
\usepackage{stfloats}
\usepackage{footmisc}
\usepackage{caption}
\usepackage{mwe}
\usepackage{subcaption}
\usepackage{thmtools}
\usepackage{thm-restate}
\usepackage{pifont}
\definecolor{icmldarkblue}{rgb}{0,0.08,0.45}
\hypersetup{
    colorlinks=true,
    linkcolor=red,
    filecolor=magenta,      
    urlcolor=blue,
    citecolor=icmldarkblue,
    pdfpagemode=FullScreen,
    }

\newcommand{\zerodisplayskips}{%
  \setlength{\abovedisplayskip}{4pt}%
  \setlength{\belowdisplayskip}{4pt}%
  \setlength{\abovedisplayshortskip}{1pt}%
  \setlength{\belowdisplayshortskip}{1pt}}
\appto{\normalsize}{\zerodisplayskips}
\appto{\small}{\zerodisplayskips}
\appto{\footnotesize}{\zerodisplayskips}

\firstpageno{1}
\begin{document}
\title{Arithmetic Control of LLMs for Diverse User Preferences:\\
Directional Preference Alignment with Multi-Objective Rewards   }
\author{\centering \name Haoxiang Wang$^{\ast 1}$~ Yong Lin$^{\ast 2}$~ Wei Xiong$^{\ast 1}$~ Rui Yang$^2$~ Shizhe Diao$^2$~ Shuang Qiu$^2$\\ Han Zhao$^1$~ Tong Zhang$^1$ \\
        \addr ~~\quad $^1$University of Illinois Urbana-Champaign\quad $^2$The Hong Kong University of Science and Technology
}

\newcommand{\customfootnotetext}[2]{{%
		\renewcommand{\thefootnote}{#1}%
		\footnotetext[0]{#2}}}%
\customfootnotetext{$\ast$}{Equal contribution. Correspondance to: Haoxiang Wang (\href{mailto:hwang264@illinois.edu}{hwang264@illinois.edu}) }

\maketitle

\vspace{0.5cm}

\begin{abstract}%
Fine-grained control over large language models (LLMs) remains a significant challenge, hindering their adaptability to diverse user needs. While Reinforcement Learning from Human Feedback (RLHF) shows promise in aligning LLMs, its reliance on \textit{scalar} rewards often limits its ability to capture diverse user preferences in real-world applications. To address this limitation, we introduce the Directional Preference Alignment (DPA) framework. Unlike the scalar-reward RLHF, DPA incorporates \textit{multi-objective reward} modeling to represent diverse preference profiles. Additionally, DPA models user preferences as \textit{directions} (i.e., unit vectors) in the reward space to achieve user-dependent preference control. Our method involves training a multi-objective reward model and then fine-tuning the LLM with a preference-conditioned variant of Rejection Sampling Finetuning (RSF), an RLHF method adopted by Llama 2. This method enjoys a better performance trade-off across various reward objectives. In comparison with the scalar-reward RLHF, DPA offers users \textit{intuitive control over LLM generation}: they can \textit{arithmetically} specify their desired trade-offs (e.g., more helpfulness with less verbosity). We also validate the effectiveness of DPA with real-world alignment experiments on Mistral-7B. Our method provides straightforward arithmetic control over the trade-off between helpfulness and verbosity while maintaining competitive performance with strong baselines such as Direct Preference Optimization (DPO). The code and trained model are released at \url{https://github.com/Haoxiang-Wang/directional-preference-alignment}.

\end{abstract}

\section{Introduction}\label{sec:intro}
Large language models (LLMs)~\citep{OpenAI2023GPT4TR, Anthropic@claude} have demonstrated remarkable capabilities across various domains and tasks, such as mathematical reasoning~\citep{wei2022chain} and medical question answering~\citep{singhal2023towards, wang2023pre, thirunavukarasu2023large}. However, for an assistant to be truly useful, it must align with human preferences, such as being helpful, honest, harmless, and managing verbosity. 

\begin{figure}[t]
\centering
    \includegraphics[width=.7\linewidth]{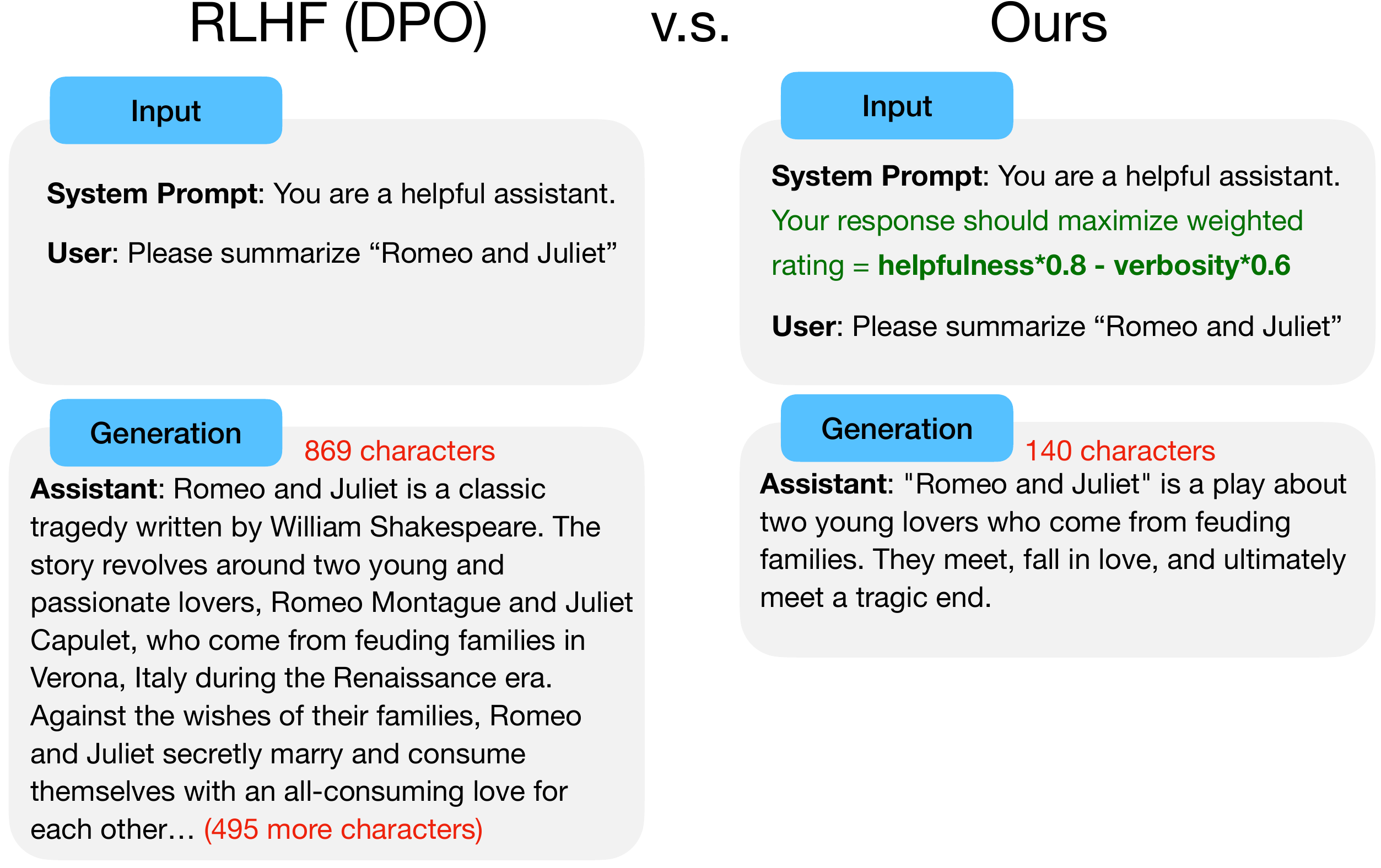}
\caption{\textbf{Arithmetic Prompting} for Preference-Conditional Generalization: Comparison between conventional RLHF methods such as DPO and our Directional Preference Alignment (DPA). In the case of DPO (left), it is capable of generating helpful responses, but these tend to be excessively verbose. Conversely, with our DPA (right), it allows for \textbf{arithmetic control} of LLMs to meet various user preferences. For instance, setting the directional preference (unit vector) to $v=\left< 0.8, -0.6\right>$ leads to less verbose responses from our aligned LLM.
}\label{fig:prompt-template}
\end{figure}

\textit{Reinforcement Learning from Human Feedback} (RLHF) \citep{christiano2017deep, ziegler2019fine,instructgpt,bai2022constitutional,lee2023rlaif}, is the leading approach to adapt LLMs towards these complex, often implicitly-defined goals. Typically, the most popular RLHF framework \citep{christiano2017deep, ziegler2019fine, instructgpt} first constructs a scalar reward model to represent the difficult-to-specify goal of being preferred by human and then use this reward model to provide signals for the subsequent reward optimization stage. Its success spans various practical applications, including recommendation systems \citep{pereira2019online}, image generation \citep{hao2022optimizing, wu2023better, dong2023raft}, robotics \citep{brown2019extrapolating}, and most notably, aligning LLMs with human values and preferences, such as ChatGPT \citep{OpenAI2023GPT4TR}, Claude \citep{Anthropic@claude}, Llama 2 \citep{llama2} and Gemini \citep{google@gemini}.

\begin{figure*}[t]
\vspace{-1em}
\centering
    \includegraphics[width=\linewidth]{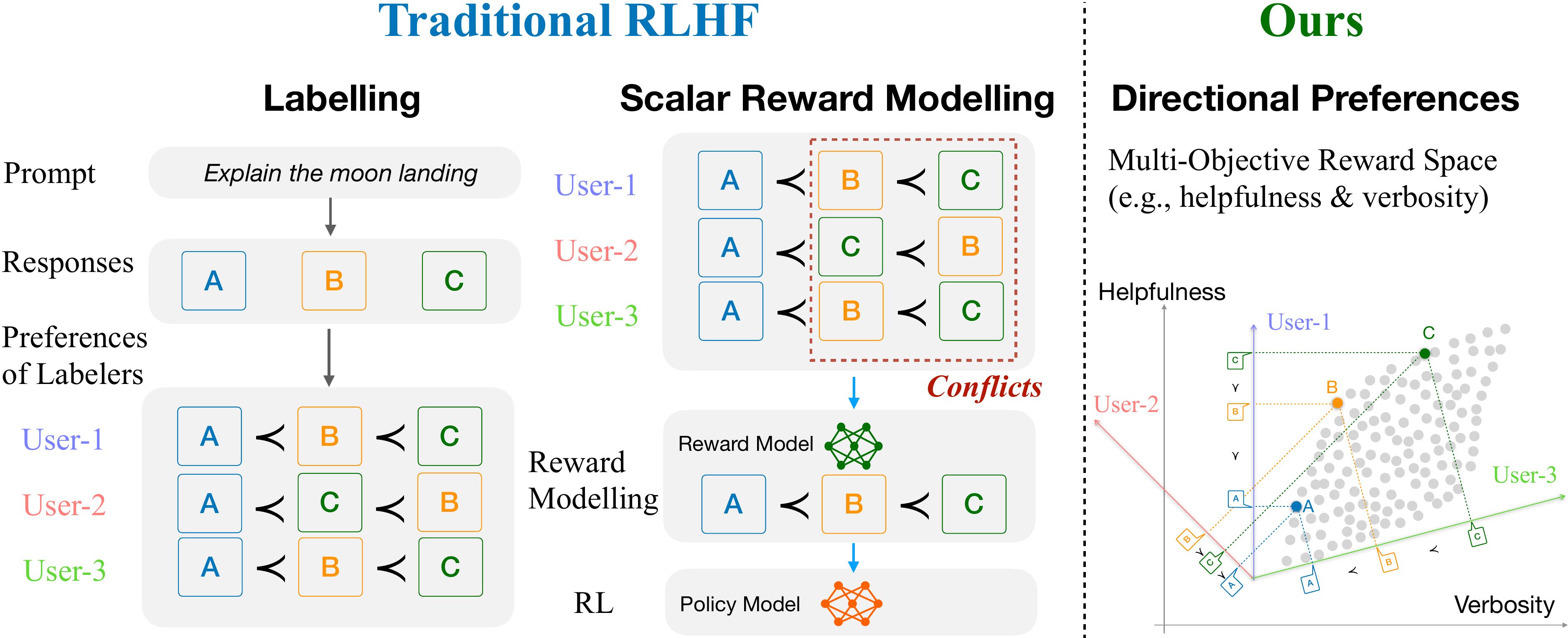}
\caption{
(Left) The illustration depicts preference conflicts among different users, where User-1 and User-3 favor response B over response C, while User-2 prefers C over B. (Middle) Generally, the scalar-reward RLHF framework tends to align toward {the average-user preference}, thus favoring B over C, which overlooks the preference of User-2. (Right) Our Directional Preference Alignment (DPA) enables users to specify their preference vector in a multi-dimensional space, allowing each user's preference to be well represented within this context.    
}\label{fig:preference-conflict}
\vspace{-1em}
\end{figure*}

While recent advancements in RLHF are noteworthy, a fundamental challenge persists due to problem misspecification. This means that a \textit{single} reward function may not sufficiently capture complex human values. For example, a generative model aligned by RLHF for helpfulness tends to produce verbose responses as shown in Figure~\ref{fig:prompt-template} (Left) \citep{singhal2023long}, even though many users prefer answers that are both helpful and concise. 
Assuming scalar-objective reward implies a \textit{total order} over preferences, which is hard to satisfy when the preference is aggregated across a diverse set of human groups \citep{may1954intransitivity, tversky1969intransitivity}, because humans typically have a set of intricate or even \emph{contradictory} targets \citep{biyik2018batch}. In real-world applications, the scalar-reward RLHF tends to align the LLMs toward an ``average-user'' preference, which cannot capture the complicated nature of human preferences and can be unfair for the under-represented groups \citep{feffer2023moral}. 
For example, consider User-1, 2, 3, and responses $A$, $B$, $C$ in Fig.~\ref{fig:preference-conflict} (Left). User-1 and 3 prefer response $B$ over $C$ ($B \prec C$), while User-2 prefers $C$ over $B$ ($C \prec B$). This could occur as response $C$ is more verbose than $B$, while User-2 prefers concise answers. When these diverse preferences are aggregated across human groups, the typical reward models with scalar rewards tend to learn the ``average-user'' preference (which is $B \prec C$ in this case), overlooking the individual preference of User-2, as shown in Figure~\ref{fig:preference-conflict} (Middle). This is also known as the ``Condorcet paradox'' in the theory of social choice \citep{gehrlein2002condorcet}. 
In general, human opinions and expertise can vary significantly \citep{coello2000handling, bobu2023aligning, bansal2023peering}. Meanwhile, the importance of these targets may also change over time, depending on the users and their expectations.

To address the limitations of the existing scalar reward model, previous works suggest the use of multi-objective rewards that characterize human preferences from different aspects (e.g., helpfulness, verbosity, harmlessness) 
\citep{pan2023rewards, rame2023rewarded}. One common way is to take the human feedback as a \textit{multi-dimensional} reward vector and each dimension models one objective \citep{rame2023rewarded,steerlm}. Then, one may apply a linear combination to transform the multi-objective rewards into a scalar for LLM alignment \citep{bakker2022fine, wu2023fine}. However, this approach still cannot handle the user-dependent needs from a diverse user population and can be unfair for minority groups. One may further adopt a user-dependent linear combination to multi-objective rewards for aligning a model for each user preference \citep{rame2023rewarded,jang2023personalized}. However, this approach is quite \emph{inference-unfriendly} because we have to switch between different models in response to the different user preferences. Finally, in social choice theory, a game-based formulation was studied under the name \textit{maximal lotteries} \citep{sternberg1965mathematics, fishburn1984probabilistic}, as well as the subsequent works in RLHF \citep{wang2023rlhf, swamy2024minimaximalist, ye2024theoretical}, to handle the diversity of user preferences. We remark that their framework is fundamentally different from the multi-objective rewards and cannot offer a user-dependent preference control in the inference stage, either. Refer to Section~\ref{sec:prior_related} for a more detailed discussion with existing methods.

In recognition of the aforementioned limitations, we propose a novel and practical alignment approach, \textit{Directional Preference Alignment} (DPA), to enhance the \emph{adaptability and controllability of a single LLM}. Our aligned LLM enjoys the flexibility to be controlled with different preferences embedded numerically into the system prompt. The ability to control preferences can significantly enhance the model's personalization ability during inference. For example, as the model is aligned with DPA with \texttt{helpfulness} and \texttt{verbosity} in consideration, a user could simply control the model's generation by specifying a directional preference $v = \left<v_1, v_2\right>$ that $\|v\|_2 = 1$, and the model will generate responses that maximize $\texttt{reward}=v_1\times\texttt{helpfulness} + v_2 \times \texttt{verbosity}$ where $\texttt{helpfulness}$ and $\texttt{verbosity}$ are rewards scored from different perspectives as shown in Figure~\ref{fig:prompt-template} (Right). Figure~\ref{fig:preference-conflict} (Right) further shows that the preferences of User-1, User-2, and User-3 can be accurately represented by specifying the preference vector in the 2-dimensional space. This is a scenario where DPA can alleviate the problem of misspecification in RLHF.   

Our approach features two crucial aspects: 1). Multi-Objective Rewards, which involve learning with multiple different preference targets simultaneously, and 2). Directional Preference Alignment, which encodes user preferences as unit vectors for preference-aware LLM alignment. Specifically, we summarize our contributions as follows.
\begin{itemize}[leftmargin=*,align=left,noitemsep,nolistsep]
    \item \textbf{We identify the limitations of existing popular RLHF frameworks}: 1) the limited capacity for capturing the real-world complicated human preference; 2) lacking in adaptability for user-dependent preference;
    \item \textbf{We propose Directional Preference Alignment (DPA)}: a novel alignment approach that allows a \textit{single} LLM to accommodate users with varying preferences.
    \item \textbf{We consider both helpfulness and verbosity rewards, and align Mistral-7B \citep{jiang2023mistral} with our DPA}: empirical evaluations show that DPA offers effective arithmetic control over the trade-off between helpfulness and verbosity, while maintaining competitive performance with DPO \citep{rafailov2023direct}.
\end{itemize}

\begin{figure*}
    \centering
    \includegraphics[width=1.0\linewidth]{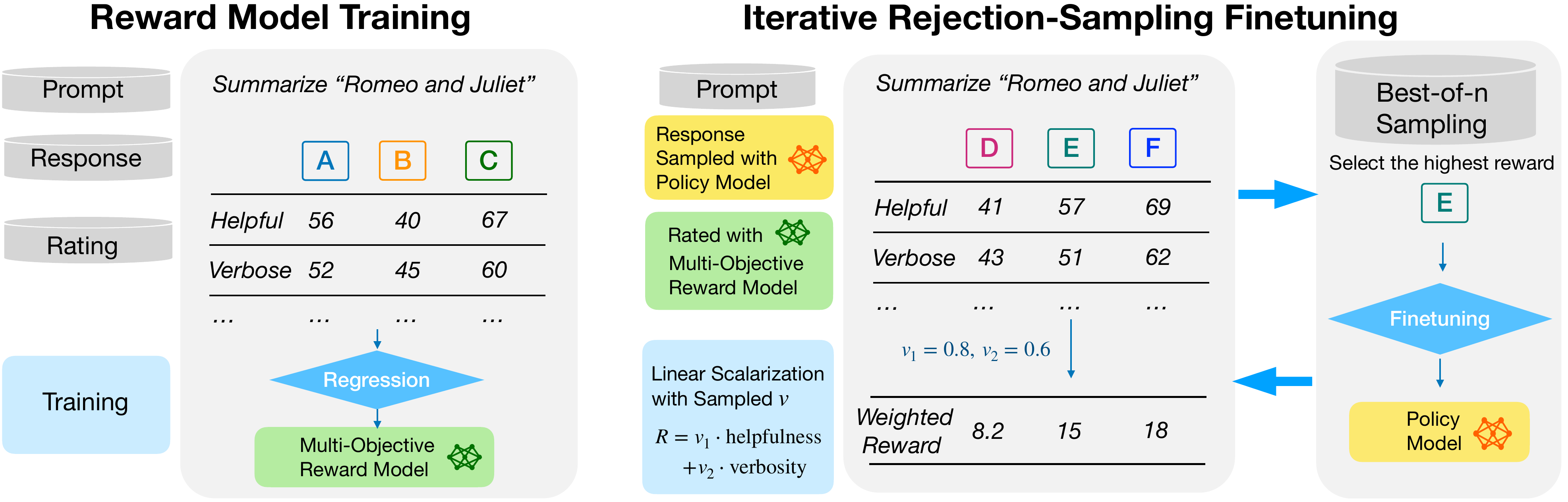}
    \caption{Illustration of the Directional Preference Alignment procedure}
    \label{fig:DPA_procedure}
    \vspace{-.8em}
\end{figure*}

\section{Directional Preference Alignment}

In a typical RLHF pipeline \citep{instructgpt, bai2022training, llama2}, we first construct a reward model based on a labeled preference dataset (e.g., preference $A\prec B \prec C$ annotated by a labeler) and then use the reward model to provide supervision for the subsequent reward optimization stage. In this section, we first present the problem setup, where we additionally consider multi-objective rewards and user preferences in the framework. Then, we present our algorithm, the Directional Preference Alignment, to handle the problem of preference-aware alignment.

\paragraph{Notation.} We denote the prompt space and the response space as $\mathcal{X}$ and $\mathcal{Y}$, respectively. $\mathbb{S}^k = \{v \in \mathbb{R}^k: \|v\|_2 = 1\}$ is the unit sphere under the $\norm{\cdot}_2$ norm. We use $\pi_\theta$ to denote the policy (generative) LLM whose parameter is $\theta$.

\subsection{Multi-Objective Reward Model}

We consider $k$-objective reward for a response $y$ given prompt $x$ as
\begin{align*}
    r(x,y) = \left<r_1(x,y), \dots, r_k(x,y)\right> \in \mathbb{R}^{k}
\end{align*}
where each $r_i(x,y)$ is the rating for a single attribute such as helpfulness, correctness, and verbosity. We use $r$ to denote $r(x,y)$ for short when it is clear from the context. Let $\mathcal{D}_r$ denote the distribution of $(x,y,r)$~\citep{helpsteer,open-assistant}.  We then train a multi-objective reward model $\tilde{r}$ with regression loss \citep{steerlm}:
\begin{align}
    \min_{\tilde r} \mathbb{E}_{(x,y,r) \sim  \mathcal{D}_r} \| \tilde r(x,y) - r(x, y)\|_2^2.
\end{align}
The trained reward model $\tilde r$ can rate any prompt-response pair $(x,y)$ across $k$ attributes.

\subsection{Directional Preference Alignment}\label{sec:method:algo}

Our work aims to learn a collection of policies that can traverse the Pareto front as efficiently as possible. Moreover, we intend to relate the learned policies to the user's preferences concerning various objectives and control the learning process according to such preferences. To make multi-objective optimization tractable and controllable, a common approach is \emph{linear scalarization} \citep{caruana1997multitask, ghane2015new, hu2023revisiting}, which takes a linear combination of multiple objectives. Through exploring all different linear combinations, the solutions to these problems can sufficiently cover a significant area of the Pareto front, which justifies the application of the linear scalarization approach.

\paragraph{Directional Preference.}  To achieve a fine-grained representation of the preference signal, we model user preference as a \emph{direction} in the multi-objective reward space, that is, a unit vector $v = \left< v_1, \dots, v_k\right> \in \mathbb{S}^k$. Then, the preference-conditioned reward is
\begin{align}
    R(x,v,y) = v^\top r(x,y) = \sum_{i=1}^k v_i r_i(x,y).
\end{align}
To incorporate user preference into the language model, we condition the text generation on \(v\) in addition to \(x\), such that the response is generated according to \(y \sim \pi_\theta(\cdot | x, v)\). For a specific $v$, the preference-conditional reward objective is 
\begin{align}\label{eq:v-cond-obj}
    J(v,\pi_\theta) = \mathbb{E}_{x\sim \mathcal{D}_x, y\sim \pi_\theta(\cdot|x,v)}[R(x,v,y)]
\end{align}
We model the directional preferences of our targeted user population as $\mathcal P_v$, a probability distribution over $\mathbb S^n$. Finally, we optimize $\theta$ by maximizing the expected reward with respect to $\mathcal P_v$:
\begin{align}\label{eq:total-obj}
    \max_{\theta} \mathbb{E}_{v\sim \mathcal P_v}\left[J(v,\pi_\theta)\right].
\end{align}

\paragraph{Reward Optimization via Rejection Sampling.} We now proceed to discuss the algorithmic designs for optimizing the RL objective in Eq.~\eqref{eq:total-obj}. While PPO is the most predominant approach for a fixed reward function \citep{OpenAI2023GPT4TR, Anthropic@claude}, it is known that PPO is unstable and sample-inefficient in aligning LLMs \citep{choshen2019weaknesses} and imposes a heavy burden on GPU memory resources \citep{instructgpt, yuan2023rrhf}. Hence, PPO requires extensive efforts to be tuned to its best performance. In light of the above limitations, we resort to an alternative approach, \emph{Rejection Sampling Fine-tuning} (RSF) \citep{dong2023raft, yuan2023rrhf, gulcehre2023reinforced}, a RLHF algorithm used in the Llama 2 project \citep{llama2}, with appealing simplicity, stability, and comparable reward gains. In essence, the original RSF learns from the best-of-$n$ policy created by the reward function. Initially, we generate $n$ responses using a base LLM and then rank them using the reward model to select the responses with the highest reward. We further finetune our LLM based on these selected samples, and this process can be repeated multiple times. 

In our scenario, to address the multi-objective nature and user-dependent preferences, we iteratively alternate among the following steps for $t=1,\dots,T$ iterations:
\begin{enumerate}
[leftmargin=*,align=left,noitemsep,nolistsep]
\setcounter{enumi}{-1}
    \item \textbf{Preparation.} Initialize an empty dataset $\mathcal D_{t}=\emptyset$. Prepare policy model $\pi_{\theta_{t-1}}$ obtained from last iteration.
    \item \textbf{Rejection Sampling.} For each randomly sampled prompt $x$ and directional preference $v$, generate $n$ responses $\{y_1,\dots y_n\}$ by $\pi_{\theta_{t-1}}(\cdot | x, v)$ and compute their multi-objective rewards by $\tilde r(x,y)$. Obtain the linear scalarization of $\tilde r(x,y)$ by $R(x,v,y_i) = v^\T \tilde r(x,y_i)$. Then, rank $y_1, ..., y_n$ according to $R(x,v,y_i)$ and select the highest-rank response $y^\star$. Add $(x,v,y^\star)$ to $\mathcal D_t$.
    \item \textbf{Finetuning.} Train on $\mathcal D_t$: $$\theta_{t} \leftarrow \argmax_\theta \mathbb{E}_{(x,v,y)\sim \mathcal D_t} [\pi_\theta(y|x,v)].$$
    
\end{enumerate}
The whole procedure of our methods is summarized in Figure~\ref{fig:DPA_procedure}.

\begin{table*}
\vspace{-1em}
    \centering
    \scriptsize
    \begin{sc}
    \resizebox{\linewidth}{!}{
    \begin{tabular}{l c c c c}
    \toprule
  Alignment Methods &  Multi-objective Rewards & Preference Arithmetic& Single Model & Feasibility Guarantee \\
     \midrule
     PPO \citep{schulman2017proximal} & \textcolor{red}{\ding{55}} & \textcolor{red}{\ding{55}} & \textcolor{green}{\ding{51}} & \textcolor{green}{\ding{51}}\\
     \midrule
DPO \citep{rafailov2023direct} & \textcolor{red}{\ding{55}} & \textcolor{red}{\ding{55}} & \textcolor{green}{\ding{51}} & \textcolor{green}{\ding{51}}\\
     \midrule
Reward Soup \citep{rame2023rewarded} & \textcolor{green}{\ding{51}} & \textcolor{green}{\ding{51}} & \textcolor{red}{\ding{55}} & \textcolor{green}{\ding{51}}\\
\midrule
SteerLM \citep{steerlm} & \textcolor{green}{\ding{51}} & \textcolor{green}{\ding{51}} &  \textcolor{green}{\ding{51}} &\textcolor{red}{\ding{55}} \\

     \midrule
Ours & \textcolor{green}{\ding{51}} & \textcolor{green}{\ding{51}} & \textcolor{green}{\ding{51}} & \textcolor{green}{\ding{51}}\\
       \bottomrule
        \end{tabular}
        }
    \end{sc}\vspace{-0.1in}
        \caption{
        Comparison among different RLHF algorithms. 
        \textbf{Multi-objective rewards}: if the algorithm considers multiple reward objectives. 
        \textbf{Preference arithmetic}: if the model allows for arithmetic control of the preference. \textbf{Single model}: if the algorithm can handle different preferences with a single LLM. \textbf{Feasibility Guarantee}: Whether the model is free from the feasibility issue that the specified control vector (prompt) could be unreachable (refer to Section~\ref{sec:prior_related} for details). }  
        \label{tab:comp}
        \vspace{-1em}
\end{table*}

\subsection{Discussion with Existing Methods} \label{sec:prior_related}

\textbf{Comparison with SteerLM~\citep{steerlm}.} 
Recall that we have multi-objective reward $r = \left< r_1, r_2, ..., r_k \right>$ of each response $y$ to the prompt $x$. \citet{steerlm} first fine-tunes the generative model to maximize the likelihood of $y$ by taking both $x$ and $r$ as the input prompts:
\begin{align*}
    \max_\theta \mathbb{E}_{(x, y, r) \sim \mathcal{D}_r} \log P_\theta(y|x, r).
\end{align*}
When presented with a new input $\bar x$, SteerLM aims to produce a response that aligns with the newly assigned multi-dimensional $\bar r$. Particularly, a user could specify $\bar r$ as ``$(\texttt{helpfulness}=10,  \texttt{verbosity}=1)"$, namely high helpfulness but low verbosity, for a new prompt $\bar x = \mbox{``Please summarize `Romeo and Juliet'"}$. SteerLM could then generate answers according to $\bar r$.
However, SteerLM will encounter a significant challenge when a user-specified $\bar r$ falls outside the feasible region of rewards for the given $\bar x$, i.e., $\bar r \notin \{r: (\bar x, y, r) \in \mathcal{D}_r\}$. In this case, if a user sets a $\bar r$ that is not achievable given $\bar x$, SteerLM may generate uncontrolled responses due to the infeasibility of $\bar r$ under $\bar x$. For example,  ``$(\texttt{helpfulness}=10,  \texttt{verbosity}=1)$" could be infeasible for $\bar x$ according to the set $\mathcal{S}$ since it will be difficult or impossible to generate a helpful summarization of `Romeo and Juliet' in very few words. 

\noindent\textbf{Comparison with Soup Methods~\citep{rame2023rewarded, jang2023personalized}.} Soup methods trains a policy $\theta_i$ for each reward objective. Let $r_i(x, y)$ denote the $i$-th objective, we have:
\begin{align*}
    \theta_i = \argmax_\theta \mathbb{E}_{x \sim \mathcal{D}_x} \mathbb{E}_{y \sim \pi_{\theta}(\cdot|x)} r_i(x, y)
\end{align*}
During inference, when a user specifies the combination vector $\left<v_1, v_2, ..., v_k\right> \in \mathbb{S}^k$, reward soups first combine the weight of $k$ models as their interpolation $\sum_i v_i \theta_i$ and then query the interpolation for response. Compared with our method, rewarded soup can cause significant storage and computation overhead because they need to maintain $k$ LLMs and calculate different interpolations whenever a new combination vector is assigned.

\vspace{-.5em}
\section{Empirical Results}

We conduct experiments on Mistral-7B \citep{jiang2023mistral}, focusing on two reward objectives: \texttt{helpfulness} and \texttt{verbosity}. Our proposed DPA achieves arithmetic control of LLM generations for different helpfulness-verbosity preferences while demonstrating an excellent balance between the two objectives.

\paragraph{Verbosity Bias.} Recently, the verbosity bias in LLMs and humans, meaning that LLMs and humans sometimes prefer more verbose answers even though they are of similar qualities, has attracted considerable attention \citep{saito2023verbosity,singhal2023long}. It has been exploited or even ``hacked'' by the RLHF-aligned models. For instance, \citet{kabir2023answers} demonstrated that $77\%$ of ChatGPT answers are verbose, while \citet{yuan2024self} found that the average output length increases to 2.5 times as the DPO iterates. Preliminary experiments have been conducted in response to this bias, such as those by \citet{chen2024odin}, which explicitly consider verbosity as a response feature. Benchmark creators like AlpacaEval \citep{alpaca_eval} and MT-Bench \citep{zheng2023judging} have observed verbosity bias in their LLM judges (typically GPT-4), and AlpacaEval-2.0 has adjusted to account for output length\footnote{\small \url{tatsu-lab.github.io/alpaca_eval/}}.

\subsection{Implementation}
\paragraph{Datasets.} We use two datasets for experiments: HelpSteer and UltraFeedback. Both datasets are used for reward model training\footnote{We include HelpSteer since it has \texttt{verbosity} annotations.}, while only UltraFeedback is used for finetuning.

\begin{itemize} [leftmargin=*,align=left,noitemsep,nolistsep]
    \item  \textbf{HelpSteer} \citet{wang2023helpsteer} comprises 10K prompts and 37K annotated responses with five attributes: \texttt{helpfulness}, \texttt{correctness}, \texttt{coherence}, \texttt{complexity}, and \texttt{verbosity}. A 43B closed-source LLM generated responses, and human labelers annotated each response on a scale of 0-4 for the five attributes.
    \item \textbf{UltraFeedback} \cite{ultrafeedback} includes 64K prompts, each of them are associated with 4 responses of five attributes: \texttt{honesty}, \texttt{truthfulness}, \texttt{instruction-following}, \texttt{helpfulness} and \texttt{overall-score}. GPT-4 was employed to label these responses. We use the same training-validation prompt split\footnote{ \url{hf.co/datasets/HuggingFaceH4/ultrafeedback_binarized}} as Zephyr \citep{tunstall2023zephyr}.
\end{itemize}     
\paragraph{Reward Modeling.} We train a multi-objective reward model on the union of HelpSteer and UltraFeedback, initializing with Mistral-7B. Specifically, we follow SteerLM-v2 practices\footnote{The authors of SteerLM \citep{steerlm} improved the original training recipe in a follow-up work \citep{helpsteer}, which we denote as SteerLM-v2.} \citep{helpsteer}, attaching a linear regression head layer on the last hidden state of Mistral-7B.  We include both regression and traditional language modeling losses in the reward model training, as we find the latter improves accuracy without additional observed costs. The reward model has 10 output dimensions: the first half corresponds to HelpSteer's five attributes, while the other half accounts for UltraFeedback's attributes. Rewards in each dimension are rescaled to the range of 0-100 in the data preprocessing stage.

\paragraph{Alignment Setup.} For a fair comparison with DPO \citep{rafailov2023direct}, we conduct a head-to-head comparison with Zephyr-$\beta$ \citep{tunstall2023zephyr}, a DPO-trained Mistral-7B model that was state-of-the-art (7B) at its release.  Zephyr-$\beta$ uses supervised fine-tuning (SFT) on UltraChat-200K \citep{ultrachat} followed by DPO on UltraFeedback \citep{ultrafeedback}. Since RLHF typically begins with SFT models, we initialize with the SFT checkpoint of Zephyr-$\beta$ and apply DPA on UltraFeedback. Following practices of \citet{ultrafeedback,tunstall2023zephyr}, we average \texttt{instruction-following}, \texttt{truthfulness}, \texttt{honesty}, and \texttt{helpfulness} ratings of UltraFeedback for the overall \textit{helpfulness} objective. We use HelpSteer's \texttt{verbosity} attribute for the verbosity objective. Our multi-objective reward model annotates \textit{helpfulness} and \textit{verbosity} for all UltraFeedback data and self-generated responses.

\paragraph{Rewards and Directional Preferences.} We denote the reward objectives for helpfulness and verbosity as $r_1$ and $r_2$, respectively. As noted by \citet{singhal2023long}, $r_1$ and $r_2$ correlate positively.  Therefore, aligning an LLM to maximize $r_1$ (helpfulness) will also tend to increase $r_2$ (verbosity), a trend documented in recent works \citep{yuan2024self,chen2024odin}. 
Consequently, when using the preference-conditional reward $v^\top r = v_1 r_1 + v_2 r_2$, we argue that it is unnecessary to have $v_2 > 0$ (i.e., to explicitly encourage verbosity). Instead, we propose sampling the distribution of $\left<v_1,v_2\right>$ as $\arctan(\frac{v2}{v1}) \sim \mathrm{Uniform}(-\frac{\pi}{4},0)$ with $v_1\in [\sqrt{2}/2,1]$ and $v_2 \in [-\sqrt{2}/2,0]$. Intuitively, this lets the user preference direction $\left<v_1,v_2\right>$ be uniformly sampled between $\left<1,0\right>$ (pure focus on helpfulness) and $\left<\sqrt{2}/2, -\sqrt{2}/2\right>$ (a balance favoring less verbosity) on the unit circle.

\paragraph{Dataset Splitting.}
Iterative RLHF methods typically sample responses for \textit{unseen} prompts in each new iteration to prevent the model from simply memorizing and repeating the responses \citep{dong2023raft, xiong2023gibbs,  yuan2024self}.  
In view of this, we split UltraFeedback dataset into two disjoint subsets, $\mathcal{D}_1$ and $\mathcal D_2$, containing an equal number of unique prompts. In each iteration $t$, we initialize the policy model $\pi_{\theta_t}$ from an SFT checkpoint rather than $\pi_{\theta_{t-1}}$, and we use a different subset from the last iteration. The use of alternative subsets ensures that the policy model $\pi_{\theta_{t}}$ for response sampling in iteration $t+1$ has not encountered the prompts before.

\paragraph{Rejection Sampling.} We conduct rejection sampling following our iterative algorithm detailed in Sec. \ref{sec:method:algo}. Notice that to launch training in $t=1$, we need $\pi_{\theta_{t=0}}$ for sampling responses for a diverse set of helpfulness-verbosity preferences. However, Zephyr-$\beta$-SFT is not designed for preference-conditional generation, making it not a good choice for $\pi_{\theta_{t=0}}$. To resolve this, we train a SteerLM model on $\mathcal D_2$ (a half of UltraFeedback) that can generate responses conditioned on both user prompt $x$ (sampled from $\mathcal D_1$) and reward objectives $r_1,r_2$. We use this model for rejection sampling in iteration $t=1$ to obtain $\pi_{\theta_{1}}$ (for each prompt, we generate 80 responses for diverse reward combinations $(r_1,r_2)$). In all the following iterations, for each prompt, we sample 5 directional preferences $\left<v_1, v_2\right>$, and use $\pi_{\theta_{t-1}}$ to generate 16 responses per preference, then keep the highest-reward response and reject the rest 15. 

\vspace{-.3em}
\paragraph{Fine-tuning.} For the response data obtained through rejection sampling, we prepend the user's directional preference to the system prompt, as illustrated in Fig.~\ref{fig:prompt-template}, to make the model aware of the user preference. The fine-tuning process then follows the same approach as SFT, optimizing the next-token prediction loss across the text corpus. It is also worth noting that RLHF often leads to performance degradation or knowledge forgetting, a phenomenon referred to as \textit{alignment tax} in the literature \citep{askell2021general, lin2023speciality}. To mitigate this issue, we adopt the memory replay techniques suggested in Instruct-GPT \citep{instructgpt} and Llama 2 \citep{llama2} that can effectively reduce alignment tax \citep{lin2023speciality}. Specifically, we incorporate original responses from UltraFeedback, which constitute about 15\% of our finetuning data for each iteration. 
Our algorithm is applied for iterations $t=1,\dots,4$. 

\paragraph{Software, Hardware and Hyperparameters} We use PyTorch \citep{pytorch} with HuggingFace's TRL framework \citep{trl} for all fine-tuning experiments across $t=0,\dots,T$. All experiments are conducted on 8x A6000 GPUs. The training cost of each DPA iteration is about 60 GPU hours. The AdamW optimizer \citep{adamw} is employed with a learning rate of $10^{-5}$ and a cosine learning rate schedule (20 warmup steps). We use a context window of 4096 tokens with sample-packing (packing short responses within the context window). The training takes 2 epochs with a global batch size of 64. We use vLLM \citep{vllm} for inference. In the rejection sampling process, we conduct inference with temperature $1.0$. In evaluation (Sec. \ref{sec:exp:eval}), we use temperature $0.7$.

\begin{figure}[t]
\centering
\vspace{-1em}
    \includegraphics[width=0.7\linewidth]{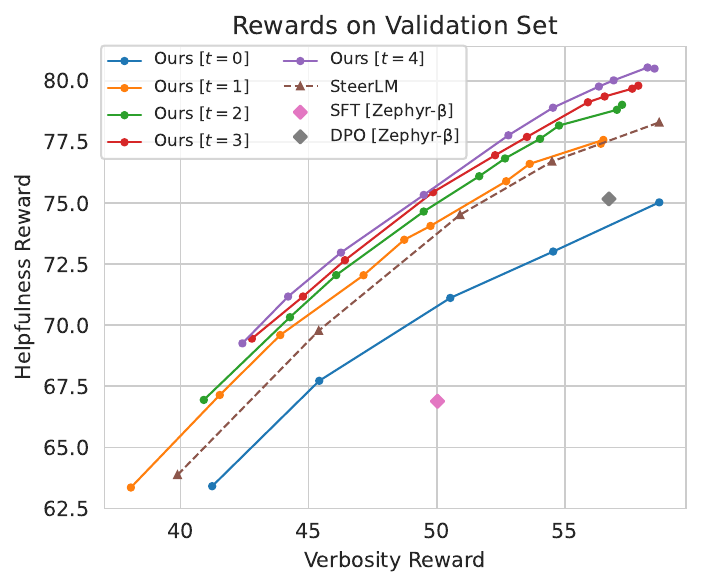}
\caption{The validation reward of different methods. When $t\geq 1$, our DPA model Pareto-dominates SFT, DPO, and SteerLM. Further, DPA at iteration $t$ Pareto-dominates models at previous iteration $t'$ with  $t' < t$. }\label{fig:validation-reward}
\vspace{-1em}
\end{figure}

\subsection{Evaluation}\label{sec:exp:eval}

\paragraph{Rewards on Validation Set} For validation, we used 2000 prompts from UltraFeedback and considered 10 uniformly sampled directional preferences ranging from \(v=\left<1,0\right>\) to \(v=\left<\sqrt{2}/2, \sqrt{2}/2\right>\). For each prompt-preference combination, our DPA-aligned models generated two responses. We then calculated the average \textit{helpfulness} and \textit{verbosity} rewards for all 2000 responses per preference using our reward model. For SteerLM\footnote{We trained a SteerLM model (initialized with the SFT checkpoint of Zephyr-$\beta$) on UltraFeedback, following practices of \citet{helpsteer}.}, five \textit{verbosity} reward values were sampled, and the highest corresponding \textit{helpfulness} reward from UltraFeedback was identified for each value. These verbosity-helpfulness pairs were then used to condition SteerLM's generation, with the average rewards computed across prompts. In the case of Zephyr-$\beta$'s DPO and SFT models, we generated responses using their original prompt templates and averaged the rewards across the validation set. The results, illustrated in Fig.~\ref{fig:validation-reward}, show that as $t\geq 1$, our DPA model Pareto-dominates SFT, DPO, SteerLM, and DPA at iteration $t$ Pareto-dominates the models of previous iterations. This demonstrates DPA's effective arithmetic control for different user preferences, and with increasing finetuning iterations \(t\), the \textit{empirical front} of DPA (i.e., each curve in Fig.~\ref{fig:validation-reward}) expands, indicating that our finetuning approach successfully maximizes rewards for all user preferences of consideration. Notably, our DPA's empirical front significantly surpasses that of SteerLM and DPO, even though all models were trained on the same UltraFeedback dataset and originated from the same SFT model.

\begin{figure}[t]
\centering
\vspace{-1em}
    \includegraphics[width=0.75\linewidth]{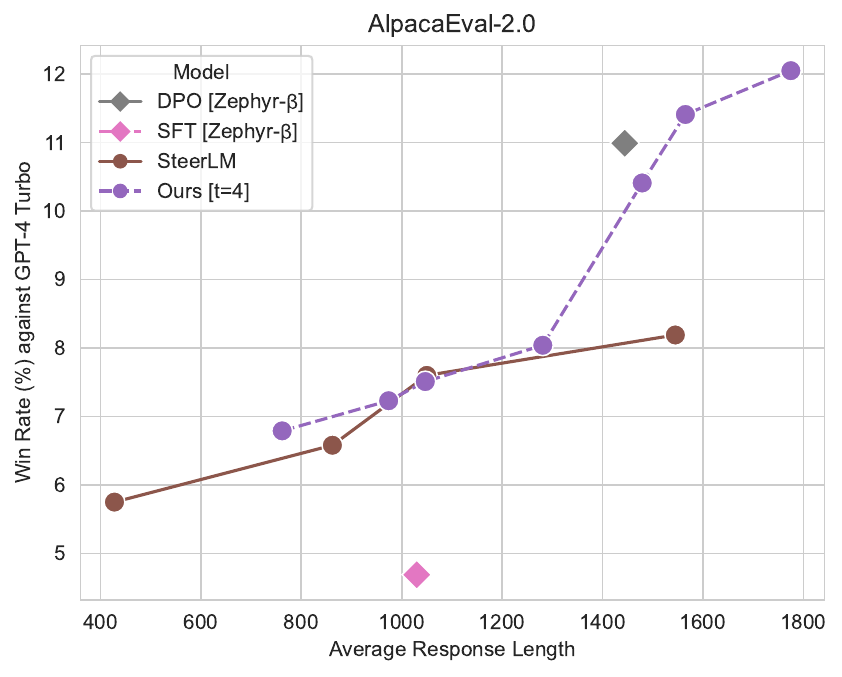}
\caption{
AlpacaEval-2.0 evaluation results.
}\label{fig:alpaca-eval}
\vspace{-1.3em}
\end{figure}
\vspace{-.3em}

\paragraph{AlpacaEval-2.0 Evaluation} AlpacaEval-2.0 \citep{alpaca_eval} is an LLM-based automatic evaluation benchmark that employs GPT-4-turbo as the LLM judge. It includes 805 prompts, and model responses to these prompts are compared with reference answers provided by GPT-4-turbo. Subsequently, the win-rate against the reference answers is calculated as a metric for the models' instruction-following capabilities. We evaluated SteerLM and our DPA (at \(t=4\)) conditioned with various user preferences and report the win rate and average response length in Fig.~\ref{fig:alpaca-eval}, along with DPO and SFT results for reference. Fig. \ref{fig:alpaca-eval} demonstrates that our DPA model outperforms SteerLM and achieves competitive performance against DPO while providing arithmetic control for diverse user preferences. The discrepancy between the validation reward evaluation results and the AlpacaEval-2.0 outcomes may arise because our reward model has different behaviors and preferences compared to GPT-4-turbo. While DPA can closely fit the reward model, this does not necessarily guarantee generalization to GPT-4-turbo evaluations.

\section{Related Works}

\paragraph{Large Language Models.} 
The landscape of natural language processing has been profoundly transformed in recent years through the development of large language models (LLMs), showcasing human-level proficiency across a range of tasks including text classification, generation, and complex reasoning. 
This progress stems from extensive pre-training on vast datasets, enabling these models to address diverse challenges.  
Despite their achievements, a distinction arises between closed-source models (e.g., GPT-3~\citep{brown2020language}, Bard \citep{google@bard}, Claude \citep{Anthropic@claude}, and PaLM~\citep{chowdhery2023palm}), often surpassing their open-source counterparts (e.g., megatron-turing-530b~\citep{smith2022using}, and Bloom~\citep{workshop2022bloom}) in performance~\citep{liang2022holistic}, which poses challenges for open-source research. 
However, initiatives like Meta's LLaMA~\citep{llama2} and subsequent works such as Alpaca~\citep{taori2023alpaca}, Vicuna~\citep{vicuna2023}, and LMFlow~\citep{diao2023lmflow}, demonstrate significant open-source contributions that continue to push the boundaries of what's possible with LLMs. 
These advancements enabled by the fine-tuning techniques, aim to improve LLMs' ability and adapt to a wide range of domains and tasks. 
Nonetheless, as these generative foundation models advance, they still face problems like implicit biases, underscoring the need for ongoing alignment and ethical considerations in their development and application. In this paper, we focus on how to align LLMs with human preferences, including the principles of being helpful, honest, and harmless as outlined by \citep{askell2021general}. This procedure is often achieved by Reinforcement Learning with Human Feedback (RLHF) \citep{instructgpt}.

\paragraph{RLHF Algorithmic Designs.} Policy Optimization (PPO) \citep{schulman2017proximal} is the most predominant approach, with its tremendous success in Chat-GPT \citep{OpenAI2023GPT4TR} and Claude \citep{Anthropic@claude}. However, PPO is significantly less efficient and stable compared to supervised finetuning \citep{choshen2019weaknesses}, and is also sensitive to the parameter and code-level implementation \citep{engstrom2020implementation}. Therefore, tuning the PPO to its best performance is very challenging in practice and the results of Chat-GPT \citep{OpenAI2023GPT4TR} have not been widely reproduced so far. In view of this, efforts have been made to develop supervised-learning-based methods as an alternative approach to the PPO, and we review them as follows. Rejection sampling finetuning (RSF) is proposed in \citep{dong2023raft, yuan2023rrhf, gulcehre2023reinforced} with different variants, but essentially, they learn from the positive samples selected by a learned reward model. RSF was applied to the RLHF of LLaMA2 project \citep{llama2} and we adopt the iterative implementation as suggested in \citet{dong2023raft, llama2, gulcehre2023reinforced}. There is also another line of work designing algorithms from the KL-constraint reward optimization \citep{rafailov2023direct, zhao2023slic, azar2023general, xiong2023gibbs}, which additionally requires the resulting model to be close to the initial model. Among them, the Direct Preference Optimization (DPO) \citep{rafailov2023direct} has attracted considerable attention due to its simplicity and stability, and effectiveness. We remark that it is also possible to incorporate these algorithmic ideas into our DPA framework and we leave the algorithmic design beyond RSF to future work.

\vspace{-.3em}
\paragraph{Fine-grained Preference Representation and Algorithmic design.} The scalar-reward-model has been criticized mainly due to its limited capacity \citep{wu2023fine, casper2023open, munos2023nash} (see the discussion of preference intransitivity in Section~\ref{sec:intro} for an illustrative example). A line of works has considered multi-objective rewards to capture the different aspects of human preferences \citep{zhou2023beyond,jang2023personalized,llama2, wu2023fine, open-assistant, rame2023rewarded}. However, the multi-objective rewards are then combined in a fixed way \citep[e.g.,][]{wu2023fine, llama2}, mainly to represent a preference averaged over different human groups, failing to capture the user-dependent preference. By introducing the user preference as a unit vector (direction) into the directional preference alignment framework, we achieve a fine-grained and user-dependent representation for the complicated human preference. Notably, in social choice theory \citep{sternberg1965mathematics, fishburn1984probabilistic}, as well as some very recent studies in RLHF \citep{wang2023rlhf, swamy2024minimaximalist, ye2024theoretical}, the RLHF is formulated as a game between two LLMs to partially handle the diversity of preferences in the population-level. The learning objective is accordingly adjusted to be solving the \textit{Nash equilibrium} of the game. In comparison, our techniques are fundamentally different from theirs and may offer computational advantages since game-based formulation is far more complicated. 

\section{Limitations}

A primary constraint of our DPA framework is its reliance on a robust multi-objective reward model. The efficacy of DPA is intrinsically linked to the precision and discriminative capability of this reward model. Should the reward model not adequately capture the subtleties of specific preferences or exhibit bias in its reward distribution, the DPA might inadvertently exacerbate these shortcomings throughout the fine-tuning process. Furthermore, if the reward model fails to recognize harmful content, it could lead the aligned model to produce such content during inference.

\section{Conclusion}

In this paper, we introduce Directional Preference Alignment (DPA) to incorporate multidimensional user preferences. DPA addresses the limitation of conventional scalar reward models by alleviating conflicting user preferences through a high-dimensional preference vector in a multidimensional space. We demonstrate that DPA efficiently explores the Pareto front in the multidimensional reward space, revealing a more effective trade-off between helpfulness and verbosity on Mistral-7B compared to existing strong baselines such as DPO.  

\newpage
\bibliography{reference}

\newpage
\appendix

\newpage

\end{document}